\pdfoutput=1

\documentclass[11pt]{article}
\usepackage[]{acl}

\usepackage{times}
\usepackage{latexsym}

\usepackage[T1]{fontenc}

\usepackage[utf8]{inputenc}

\usepackage{microtype}

\usepackage{inconsolata}

\usepackage{graphicx}
\usepackage{tabularx} 
\usepackage{inconsolata}
\usepackage{pifont}
\usepackage{algorithm}
\usepackage{algorithmic}
\usepackage[namelimits]{amsmath} 
\usepackage{amssymb}             
\usepackage{amsfonts}            
\usepackage{mathrsfs}            
\usepackage{graphicx} 
\usepackage{subfigure} 
\usepackage{xcolor}
\definecolor{darkgreen}{HTML}{3CB371}
\definecolor{darkred}{HTML}{7B7B7B}
\definecolor{ggreen}{HTML}{d4e7cf}
\definecolor{bblue}{HTML}{00bfff}
\definecolor{ppink}{HTML}{ff69b4}

\usepackage{makecell}
\usepackage{booktabs}
\usepackage{multirow}
\usepackage{enumitem} 
\usepackage{CJKutf8}

\usepackage{newfloat}
\usepackage{listings}
\usepackage{subcaption}
\usepackage{xspace}
\usepackage{graphicx}
\usepackage{caption}

\usepackage{amsmath}
\usepackage{mathtools}

\definecolor{green}{RGB}{36, 214, 36}
\definecolor{red}{RGB}{235, 30, 30}
\usepackage[utf8]{inputenc}
\usepackage[T1]{fontenc}

\usepackage{CJKutf8}
\usepackage{booktabs} 

\usepackage{CJK}
\usepackage{xcolor}
\usepackage{ulem}

\usepackage{booktabs}
\usepackage{tcolorbox}

\usepackage{soul}
\newcommand{\kaiti}[1]{\begin{CJK*}{UTF8}{gkai} #1 \end{CJK*}}
\soulregister{\kaiti}7
\definecolor{MyYellow}{rgb}{254, 246, 170}
\definecolor{MyBlue}{rgb}{170, 217, 251}

\usepackage{color}
\usepackage{colortbl}
\definecolor{g1}{RGB}{232,232,232}
\definecolor{g2}{RGB}{207,207,207}
\definecolor{g3}{RGB}{181,181,181}
\definecolor{g4}{RGB}{156,156,156}

\definecolor{b1}{RGB}{217,217,254}
\definecolor{b2}{RGB}{198,198,253}
\definecolor{b3}{RGB}{180,180,252}
\definecolor{b4}{RGB}{162,162,252}
\definecolor{b5}{RGB}{136,136,255}

\definecolor{r1}{RGB}{254,236,236}
\definecolor{r2}{RGB}{254,217,217}
\definecolor{r3}{RGB}{253,198,198}
\definecolor{r4}{RGB}{253,180,180}
\definecolor{r5}{RGB}{252,128,127}

\definecolor{ack}{RGB}{96,169,23}
\definecolor{rel}{RGB}{27,161,226}
\definecolor{app}{RGB}{255,128,0}
\definecolor{rea}{RGB}{102,0,204}
\definecolor{ans}{RGB}{155,0,0}

\title{Robust Knowledge Editing via Explicit Reasoning Chains for Distractor-Resilient Multi-Hop QA}

\author{%
  Yuchen Wu$^{1}$,
  Liang Ding$^{2}$\thanks{Correspond to Liang Ding \texttt{liangding.liam@gmail.com}},
  Li Shen$^{3}$,
  Dacheng Tao$^{4}$\\
  $^{1}$Shanghai Jiao Tong University, China 200240\\
  $^{2}$The University of Sydney, Australia 2006\\
  $^{3}$Shenzhen Campus of Sun Yat-sen University, China 518107\\
  $^{4}$Nanyang Technological University, Singapore 639798 
  }

\begin{document}
\maketitle
\begin{abstract}
Large language models (LLMs) encode vast amounts of world knowledge but remain static once trained, making the timely integration of emerging facts prohibitively expensive via full retraining. Knowledge‐editing techniques have thus emerged to inject or overwrite specific facts into LLMs, yet they either over‐rely on superficial cues or incur complex, iterative pipelines that collapse under noisy, multi‐hop conditions. We introduce \textbf{Reason-KE}, an \textit{end‐to‐end reasoning-chain-based editing framework} that steers a pretrained LLM through four structured stages—fact acknowledgment, relevance determination, selective application, and final reasoning—to filter distractors \textit{in a single pass}. Trained on MQuAKE‐CF with up to four irrelevant facts, Reason-KE elevates Qwen2.5‐7B’s multi‐hop QA accuracy to 90.2\% (↑17.6 pp) while suffering merely a 6.3\% drop under heavy distraction and <1\% when answers are leaked. Our quantitative analysis confirms Reason-KE’s resilience and efficiency, establishing a new state-of-the-art for reliable LLM knowledge updates. Our code is available at: \url{https://github.com/YukinoshitaKaren/Reason-KE}.
\end{abstract}

\section{Introduction}

\begin{figure}[t]
\centering
\includegraphics[width=1.0\columnwidth]{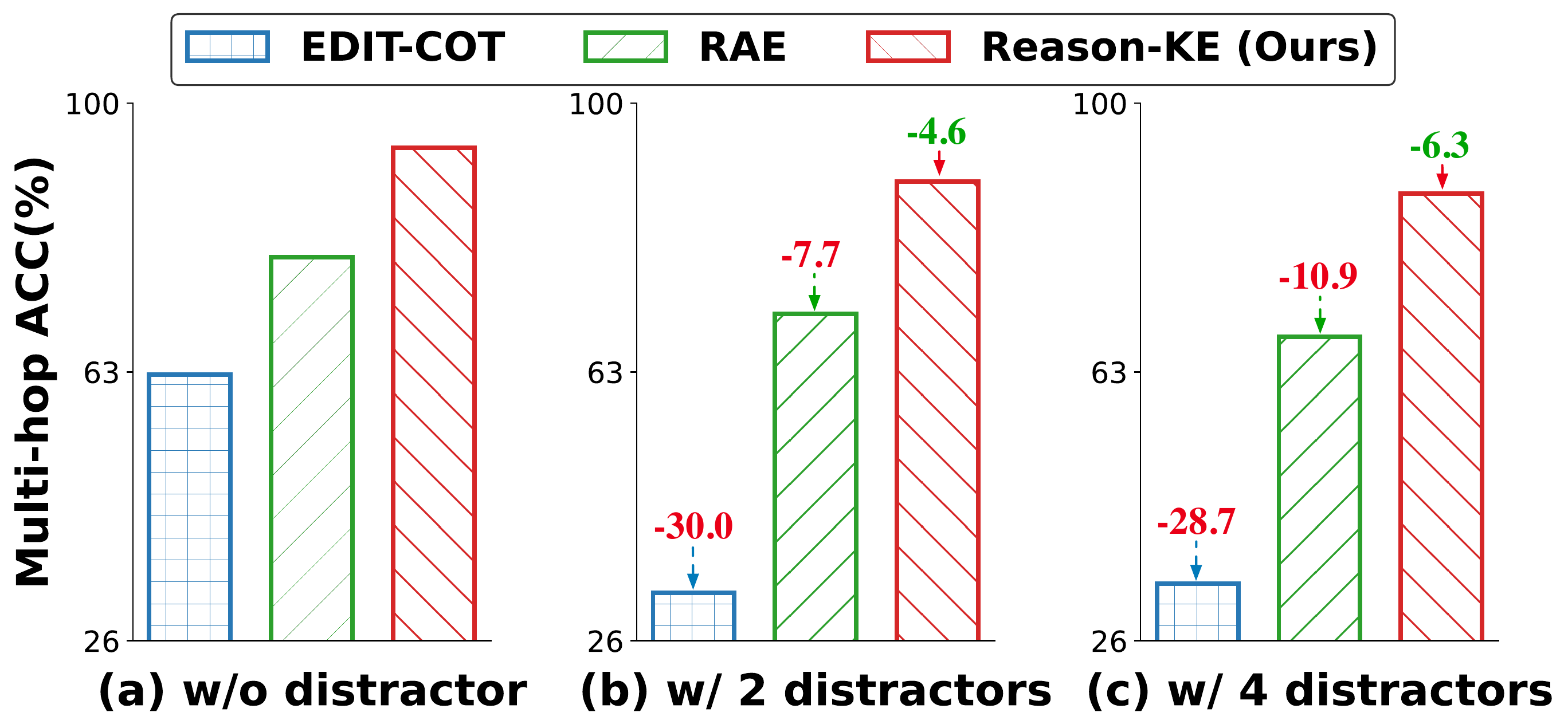}
  \caption{
  \textbf{Performance of three knowledge-editing methods under increasing levels of distraction}.
(a) w/o distractor: Editing sets contain only relevant facts.
(b) w/ 2 distractors: Two irrelevant facts are added for each relevant fact.
(c) w/ 4 distractors: Four irrelevant facts are added for each relevant fact.
Numeric annotations above each bar indicate the relative accuracy drop (↓) from the no-distractor setting. Our method consistently achieves the highest accuracy and exhibits the smallest degradation as the number of distractors grows.}
  \label{fig:figure1}
\end{figure}

Large language models (LLMs,~\citealp[]{grattafiori2024llama,qwen25,deepseek}) have shown strong capabilities in natural language understanding, generation, and reasoning~\cite{wei2022emergent, zhong2023can, peng2023towards, survey}. However, these models encode world knowledge statistically, and updating emerging facts via full retraining is prohibitively expensive. To address this limitation, knowledge editing (KE) techniques~\cite{editing,modifying,rome,memit,EDITCOT,ice,wu2025edit} have been proposed to inject or overwrite the specific facts in pretrained LLMs without retraining from scratch.

Existing KE methods can be categorized into two main groups. Parameter modification approaches~\cite{modifying,rome,memit} directly alter model weights to integrate new information, while parameter preservation methods~\cite{EDITCOT,ice,wu2025edit} add lightweight modules or leverage in-context learning to achieve editing with minimal changes to the base model. Although parameter-preservation frameworks perform well on multi-hop question answering (MQA) benchmarks~\cite{mquake,ripple}, they often rely too heavily on surface-level context cues. As a result, their performance degrades sharply when faced with noisy or irrelevant facts, a scenario common in real-world applications~\cite{deepedit}.

\begin{figure*}[!t]
\centering
\includegraphics[width=1\linewidth]{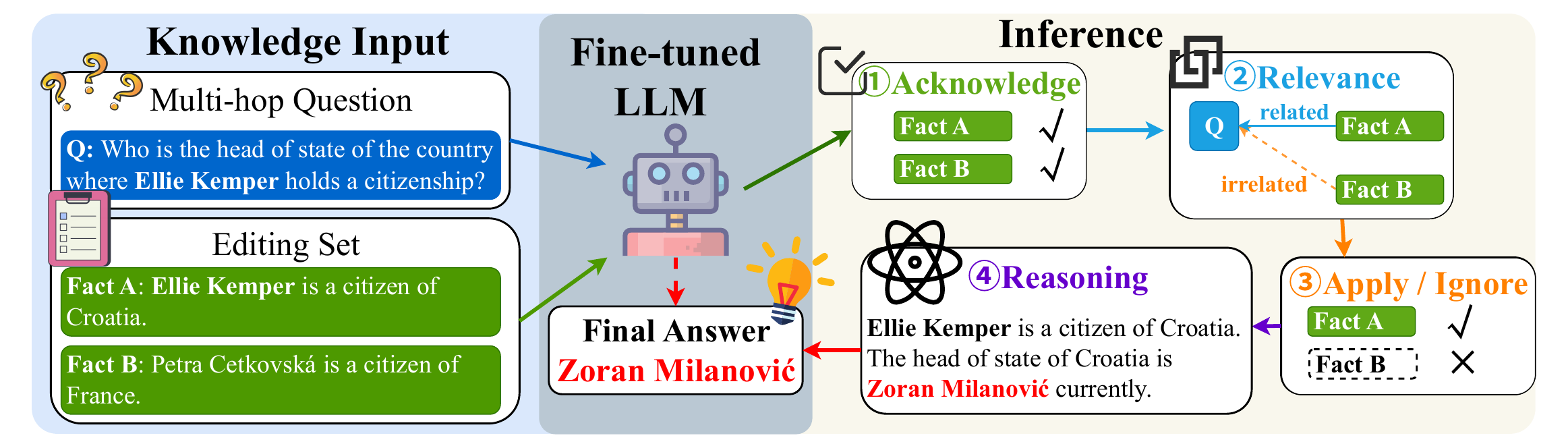}
\caption{\textbf{Illustration of our four-stage workflow for multi-hop KE in the presence of distractors.}
1. {\color{ack}\textit{Acknowledge Updated Information}} ingests all editing facts (``Fact A \& B'').
2. {\color{rel}\textit{Determine Relevance}} evaluates each fact against the question to gauge usefulness.
3. {\color{app}\textit{Apply or Ignore}} retains relevant facts (``Fact A''), discards others (``Fact B'').
4. {\color{rea}\textit{Reasoning}} composes an explicit reasoning chain to derive the {\color{red}final answer} (``Zoran Milanović'').
This pipeline filters out noise and enables reliable multi‐hop reasoning even in the presence of redundant information.}
\label{fig:method}
\end{figure*}

To overcome these challenges, we introduce \textbf{Reason-KE}, an end-to-end reasoning-chain-based KE framework that guides a pretrained LLM through four structured stages: 1) \textit{Acknowledgment} of updated information; 2) \textit{Relevance Determination} to filter distractors; 3) \textit{Selective Application} of pertinent facts; and 4) Final \textit{Reasoning} to derive the answer. By explicitly modeling each reasoning step in a single pass\footnote{Previous studies on multi-step math reasoning~\cite{zhong2024achieving} and jailbreak defense~\cite{zhang2025intention} of LLMs have shown the effectiveness of merging multiple steps into one pass.}, Reason-KE eliminates the need for complex iterative pipelines and maintains robustness under heavy distraction (Figure~\ref{fig:figure1}).

Our experiments on the MQuAKE-CF dataset~\cite{mquake}, augmented with up to four irrelevant facts, show that our proposed Reason-KE method enhances Qwen2.5-7B~\cite{qwen25}'s multi-hop QA accuracy to 90.2\%, a 17.6 percentage-point gain over the strongest baseline, while limiting performance drops to 6.3\% under high distraction and below 1\% when answers are directly exposed. 

Our \textbf{contributions} are threefold:
\begin{itemize}
    \item We propose Reason-KE, a simple and effective end-to-end framework for multi-hop knowledge editing that robustly handles redundant information without iterative loops.
    \item We empirically demonstrate that our method consistently outperforms state-of-the-art (SOTA) baselines across diverse distractor settings and exhibits minimal degradation when answers are leaked.
    \item We provide quantitative analyses on the impact of distractors and answer exposure, establishing a new SOTA in both reliability and efficiency for LLM knowledge updates.
\end{itemize}

\section{Methodology}

\subsection{Background}
\paragraph{Knowledge Editing.} 
Knowledge editing (KE) seeks to update specific facts in a pretrained LLM without full retraining~\cite{fast}.  Each fact is a triplet $f = (s, r, o)$, and an edit changes the object to $o^*$, yielding $e = (s, r, o \rightarrow o^*)$, for example, (\textit{the United States, the president of \{ \} is, Joe Biden $\rightarrow$ Donald Trump}).

\paragraph{Multi-hop QA in KE.}  Multi-hop questions $Q$ require reasoning over a chain of interdependent facts $C=[(s_1,r_1,o_1),\dots,(s_n,r_n,o_n)]$, where $s_{i+1}=o_i$ and $o_n$ is the answer.  Under editing, any $e\in\mathcal{E}$ can alter the final result. Prior work finds the golden chain $C^*$ via iterative decomposition~\cite{EDITCOT,mquake,pokemqa}, but demands highly relevant facts. 
In realistic settings, editing sets often include distractors, which may confuse the model, underscoring the need for noise resistance.

\subsection{Reasoning Training}

To mitigate the issue above, We propose Reason-KE, a simple yet efficient end-to-end framework. Unlike previous methods, Reason-KE eliminates complex iterative framework and instead explores the model’s intrinsic reasoning ability. 
Specifically, given an editing set $\mathcal{E}$ and a corresponding question $Q$, the LLM is required to generate an explicit reasoning process that leads to the final answer.

\paragraph{Construction of Training Data.}

We used the widely used dataset MQuAKE-CF in the field of knowledge editing, which contains 9,218 data points, to construct our training dataset. 
Specifically, given an editing set $\mathcal{E}$ and its corresponding QA pair, we employ Deepseed-R1\footnote{\url{https://api-docs.deepseek.com/}} to generate a step-by-step reasoning process.
Moreover, to enhance the model's robustness to distractors, we add extra 0, 2, and 4 distractors to editing sets $\mathcal{E}$ at a ratio of 90\%, 5\%, and 5\%, respectively. 
To achieve our goal, our designed reasoning process includes the following parts: 
(1) \textbf{Acknowledge Updated Information:} Confirm the facts in $\mathcal{E}$.
(2) \textbf{Determine Relevance:} Determine the relevance between the facts and question $Q$.
(3) \textbf{Apply Updated Information or Ignore:} Based on the relevance in (2), determine which fact to apply or ignore.
(4) \textbf{Reasoning:} Based on the applied knowledge and $Q$, reason out the final answer. Further details are provided in Appendix~\ref{sec:Dataset_sample}.

\paragraph{Fine-Tuning.}
We conducted the main experiments on Qwen2.5-7B-Instruct~\cite{qwen25}, and more training details can be found in Appendix~\ref{sec:appendix_implementation}. Note that our aim is to validate whether the four explicit knowledge-editing stages can be intrinsically trained for one-pass long-CoT reasoning, thus, we only adopt vanilla fine-tuning, and reinforcement-learning supervision of each intermediate step will be explored in future work.

\section{Experimental Setup}
\paragraph{Baselines and Models.}
We compare our method with various model editing methods, including the parameter modification method ROME~\cite{rome} and parameter preservation methods MeLLo~\cite{mquake}, PokeMQA~\cite{pokemqa}, EditCoT~\cite{EDITCOT}, and RAE~\cite{rae}. 
For MeLLo and PokeMQA, we employ deepseek-v3~\cite{deepseekv3}\footnote{Our preliminary study showed that training-free methods, e.g., MeLLo and PokeMQA, yielded unsatisfactory results with the Qwen model. As the performance of these methods is highly dependent on the base model's inherent capabilities, we used the more powerful Deepseek-v3 API to ensure a fair evaluation of their potential.} as the backbone models, and the other are conducted on Qwen2.5-instruct-7B~\cite{qwen25}. In addition to the Qwen model, we also validate the effectiveness of our method in Llama3-8B-Instruct~\cite{grattafiori2024llama}. More details are provided in Appendix~\ref{sec:details_baselines}.

\paragraph{Datasets and Metrics.}
We evaluate our method and baselines on the MQuAKE dataset~\cite{mquake}, a knowledge editing benchmark designed for multi-hop QA. We use MQuAKE-CF-3k as our test set, which includes 3,000 items.
Importantly, MQuAKE-CF-3k and MQuAKE-CF share no overlapping data points.
Following previous work~\cite{mquake}, we choose \textit{Multihop-Accuracy} as the evaluation metric, which uses Exact Match to measure the accuracy. More details can be found in Appendix~\ref{sec:appendix_Datasets}.

\paragraph{Distractors Selection.}
To systematically evaluate robustness to irrelevant-fact interference, we retrieve $k$ extra facts for each supporting fact needed by the question, where $k\in \{0,1,2\}$. Consequently, if a question requires $m$ facts, the total number of distractors added to $\mathcal{E} $ is $ n = m\times k  $. See Appendix~\ref{sec:selection} for more details.

\begin{table*}[t]
    \centering
    \resizebox{1.0\linewidth}{!}{%
    \begin{tabular}{lcccccccccc}
        \toprule
        \multirow{2}{*}{\bf Method}
            & \multicolumn{3}{c}{\textbf{2-hops}}
            & \multicolumn{3}{c}{\textbf{3-hops}}
            & \multicolumn{3}{c}{\textbf{4-hops}}
            & \multirow{2}{*}{\textit{\textbf{Avg.}}} \\
        \cmidrule(lr){2-4}\cmidrule(lr){5-7}\cmidrule(lr){8-10}
        & \textbf{w/o Distr.} & \textbf{w/ 2 Distr.} & \textbf{w/ 4 Distr.}
        & \textbf{w/o Distr.} & \textbf{w/ 2 Distr.} & \textbf{w/ 4 Distr.}
        & \textbf{w/o Distr.} & \textbf{w/ 2 Distr.} & \textbf{w/ 4 Distr.}
        & \\  
        \midrule
        ROME & \underline{12.00} & 12.00 & 11.99{\scriptsize\color{green}$\downarrow$}
            & \underline{8.83} & 8.95 & 9.11
            & \underline{5.46} & 5.68 & 5.50
            & 8.84 \\ 
        Mello & \underline{80.90} & 70.90{\scriptsize\color{red}$\downarrow$} & 65.80{\scriptsize\color{red}$\downarrow\downarrow$}
            & \underline{40.30} & 29.50{\scriptsize\color{red}$\downarrow$} & 30.40{\scriptsize\color{red}$\downarrow\downarrow$}
            & \underline{9.30} & 10.40 & 11.00
            & 38.72 \\  
        PokeMQA & \underline{84.10} & 77.80{\scriptsize\color{red}$\downarrow$} & 78.30{\scriptsize\color{green}$\downarrow$}
            & \underline{61.40} & 50.90{\scriptsize\color{red}$\downarrow$} & 49.40{\scriptsize\color{red}$\downarrow\downarrow$}
            & \underline{16.00} & 12.70{\scriptsize\color{green}$\downarrow$} & 9.10{\scriptsize\color{red}$\downarrow$}
            & 48.86 \\
            
        EditCoT & \underline{76.40} & 51.80{\scriptsize\color{red}$\downarrow\downarrow$} & 54.70{\scriptsize\color{red}$\downarrow\downarrow$}
            & \underline{44.00} & 16.10{\scriptsize\color{red}$\downarrow\downarrow$} & 16.90{\scriptsize\color{red}$\downarrow\downarrow$}
            & \underline{67.50} & 30.00{\scriptsize\color{red}$\downarrow\downarrow$} & 30.10{\scriptsize\color{red}$\downarrow\downarrow$}
            & 43.06 \\
        RAE & \underline{88.90} & 87.50{\scriptsize\color{green}$\downarrow$} & 85.30{\scriptsize\color{green}$\downarrow$}
            & \underline{71.10} & 60.10{\scriptsize\color{red}$\downarrow$} & 58.10{\scriptsize\color{red}$\downarrow\downarrow$}
            & \underline{76.30} & 65.50{\scriptsize\color{red}$\downarrow$} & 60.20{\scriptsize\color{red}$\downarrow\downarrow$}
            & 72.56 \\
        \cmidrule{1-11}
        Reason-KE & \underline{\textbf{97.00}} & \textbf{96.70}{\scriptsize\scriptsize\color{green}$\downarrow$} & \textbf{96.70}{\scriptsize\color{green}$\downarrow$}
            & \underline{\textbf{88.90}} & \textbf{85.20}{\scriptsize\color{green}$\downarrow$} & \textbf{84.80}{\scriptsize\color{green}$\downarrow$}
            & \underline{\textbf{95.60}} & \textbf{85.80}{\scriptsize\color{red}$\downarrow$} & \textbf{81.10}{\scriptsize\color{red}$\downarrow$}
            & \textbf{90.20} \\        
        \bottomrule
    \end{tabular}}
    \caption{\textbf{Multi-hop performance} with \textbf{bolded} best results. Baseline performance (0 distractors) is underlined in \textbf{w/o Distr.} columns, with interference impacts quantified in \textbf{w/ 2/4 Distr.} columns. 
    {\scriptsize\color{red}$\downarrow$} indicates \textgreater6\% performance drop from w/o Distr., 
    {\scriptsize\color{red}$\downarrow\downarrow$} denotes \textgreater12\% catastrophic drop, and 
    {\scriptsize\color{green}$\downarrow$} shows stable performance (\textless6\%).}
    \label{table:hops}
\end{table*}

\begin{table*}[t]
    \centering
    \resizebox{1.0\linewidth}{!}{%
    \begin{tabular}{lcccccccccc}
        \toprule
        \multirow{2}{*}{\bf Method}
            & \multicolumn{3}{c}{\textbf{\#Edits: 1}}
            & \multicolumn{3}{c}{\textbf{\#Edits: 2}}
            & \multicolumn{3}{c}{\textbf{\#Edits: 3 \& 4}}
            & \multirow{2}{*}{\textit{\textbf{Avg.}}} \\
        \cmidrule(lr){2-4}\cmidrule(lr){5-7}\cmidrule(lr){8-10}
        & \textbf{w/o Distr.} & \textbf{w/ 2 Distr.} & \textbf{w/ 4 Distr.}
        & \textbf{w/o Distr.} & \textbf{w/ 2 Distr.} & \textbf{w/ 4 Distr.}
        & \textbf{w/o Distr.} & \textbf{w/ 2 Distr.} & \textbf{w/ 4 Distr.}
        & \\  
        \midrule
        ROME & \underline{9.36}  & 9.37  & 9.47 
            & \underline{9.81}  & 9.97  & 9.97
            & \underline{6.66}  & 6.85  & 6.70
            & 8.68 \\
        Mello & \underline{41.54}  & 35.41{\scriptsize\color{red}$\downarrow$} & 32.11{\scriptsize\color{red}$\downarrow$}
            & \underline{55.67}  & 48.83{\scriptsize\color{red}$\downarrow$} & 47.70{\scriptsize\color{red}$\downarrow$ }
            & \underline{30.60}  & 23.81{\scriptsize\color{red}$\downarrow$} & 25.24{\scriptsize\color{red}$\downarrow$}
            & 37.88 \\     
        PokeMQA & \underline{59.38}  & 54.35{\scriptsize\color{green}$\downarrow$} & 53.34{\scriptsize\color{red}$\downarrow$}
            & \underline{63.92}  & 56.23{\scriptsize\color{red}$\downarrow$} & 55.48{\scriptsize\color{red}$\downarrow$}
            & \underline{33.81}  & 26.19{\scriptsize\color{red}$\downarrow$} & 22.98{\scriptsize\color{red}$\downarrow$}
            & 47.30 \\     
        EditCoT & \underline{64.59} & 49.86{\scriptsize\color{red}$\downarrow\downarrow$} & 49.13{\scriptsize\color{red}$\downarrow\downarrow$}
            & \underline{64.57} & 35.52{\scriptsize\color{red}$\downarrow\downarrow$} & 39.46{\scriptsize\color{red}$\downarrow\downarrow$}
            & \underline{57.62} & 6.55{\scriptsize\color{red}$\downarrow\downarrow$} & 7.02{\scriptsize\color{red}$\downarrow\downarrow$}
            & 41.59 \\
        RAE & \underline{65.97} & 63.04{\scriptsize\color{green}$\downarrow$} & 60.11{\scriptsize\color{green}$\downarrow$}
            & \underline{81.07} & 68.98{\scriptsize\color{red}$\downarrow\downarrow$} & 67.39{\scriptsize\color{red}$\downarrow\downarrow$}
            & \underline{92.50} & 84.05{\scriptsize\color{red}$\downarrow$} & 78.57{\scriptsize\color{red}$\downarrow\downarrow$}
            & 73.52 \\  
        \cmidrule{1-11}
        Reason-KE & \underline{\textbf{89.84}} & \textbf{84.08}{\scriptsize\color{green}$\downarrow$} & \textbf{84.26}{\scriptsize\color{green}$\downarrow$}
            & \underline{\textbf{97.00}} & \textbf{90.25}{\scriptsize\color{red}$\downarrow$} & \textbf{85.85}{\scriptsize\color{red}$\downarrow$}
            & \underline{\textbf{95.00}} & \textbf{94.64}{\scriptsize\color{green}$\downarrow$} & \textbf{93.93}{ \scriptsize\color{green}$\downarrow$}
            & \textbf{90.54} \\
        \bottomrule
    \end{tabular}}
    \caption{\textbf{Multi-edit performance} with best results \textbf{bolded}. All markers follow the same conventions as Table~\ref{table:hops}.}
    \label{table:edits}
\end{table*}

\begin{table}[ht]
    \centering
    \resizebox{1.0\linewidth}{!}{%
    \begin{tabular}{lccc}
        \toprule
        \multirow{2}{*}{\bf Method}
            & \multicolumn{3}{c}{\textbf{Answer w/ exposed}} \\
        \cmidrule(lr){2-4}
        & \textbf{w/o Distr.} & \textbf{w/ 2 Distr.} & \textbf{w/ 4 Distr.} \\
        \midrule
        Mello & \underline{56.75} 
            & 48.06 ({\scriptsize \textcolor{red}{$\downarrow$8.69}}) 
            & 46.54 ({\scriptsize \textcolor{red}{$\downarrow$10.2}}) \\
        PokeMQA & \underline{60.21}
            & 51.30 ({\scriptsize \textcolor{red}{$\downarrow$8.91}}) 
            & 50.27 ({\scriptsize \textcolor{red}{$\downarrow$9.94}}) \\
        EditCoT & \underline{64.25} 
            & 25.49 ({\scriptsize \textcolor{red}{$\downarrow$38.8}}) 
            & 27.32 ({\scriptsize \textcolor{red}{$\downarrow$36.9}}) \\
        RAE & \underline{94.98} 
            & 88.82 ({\scriptsize \textcolor{red}{$\downarrow$6.16}}) 
            & 85.15 ({\scriptsize \textcolor{red}{$\downarrow$9.83}}) \\
        \cmidrule{1-4}
        Reason-KE & \underline{\textbf{97.08}} 
            & \textbf{96.70} ({\scriptsize \textcolor{green}{$\downarrow$0.38}}) 
            & \textbf{96.71} ({\scriptsize \textcolor{green}{$\downarrow$0.37}}) \\
        \bottomrule
    \end{tabular}}
    \caption{\textbf{Performance in the answer-exposed setting}, where {\scriptsize \textcolor{red}{$\downarrow$}} indicate \textgreater5\% degradation from w/o Distr. and {\scriptsize \textcolor{green}{$\downarrow$}} arrows denote \textless1\%.}
    \label{table:expose_ablate}
\end{table}

\section{Main Rsults}
We demonstrate the result of our method and baselines in Table~\ref{table:hops}\&~\ref{table:edits}. We demonstrate that:

\paragraph{Reason-KE outperforms other methods in multi-hop setting by a significant margin.} 

As shown in Table~\ref{table:hops}, our method achieves a 17.64\% average performance improvement over RAE in multi-hop settings, demonstrating strong reasoning capability.
Moreover, when encountering distractors, prior methods like PokeMQA and EDIT-COT struggle to construct the golden path. In contrast, our method enables the model to reason internally, preserving consistency throughout the reasoning chain. Even under the most challenging 4-hops setting, our method still maintains high performance.

\paragraph{Reason-KE brings stable improvements on more complex editing sets $\mathcal{E}$.} 
The more facts that need editing, the more interference terms added and the more complex the editing set $\mathcal{E}$ becomes.
This setting demands that the model not only comprehends the facts in $\mathcal{E}$ but also exhibits robust reasoning capabilities to arrive at the final answer.
As demonstrated in Table~\ref{table:edits}, most existing methods exhibit significant performance degradation in multi-edit scenarios (number of edits > 1). Although RAE employs pruning strategies to filter some redundant information, it does not enhance the model’s fundamental capability, resulting in suboptimal performance.
In contrast, our method can independently identify which facts are relevant to the question. This enables it to effectively filter out distractors even in complex multi-edit conditions (number of edits > 2), achieving the best performance.

\paragraph{Reason-KE does not rely on the leakage of answers in facts.}
\label{sec:leakage}

In our analysis of the MQUAKE-CF-3K dataset, we found that in 1,852 instances, the object $o^*$ of the fact triple $(s, r, o^*)$ aligns exactly with the final answer $o^*_n$ of the multi-hop question $Q$. 
This observation raises concerns about potential shortcut learning, whereby models may obtain answers directly from these fact triples without truly comprehending the updated information. To investigate, we isolate instances where answers are directly exposed and conduct experiments.
As Table~\ref{table:expose_ablate} shows, when introducing distractors to $\mathcal{E}$ under leakage conditions, most methods' performance drops sharply, revealing their dependence on superficial pattern matching. However, our method exhibits a performance drop of less than 1\%, proving its independence from surface-level editing.

\section{Analysis}

\begin{table}[t]
    \centering
    \resizebox*{1.0\linewidth}{!}{%
    \begin{tabular}{lcccc}
        \toprule
        \multirow{2}{*}{\bf Method} 
            & \multicolumn{3}{c}{\textbf{Multi-hop acc}} 
            & \multirow{2}{*}{\textit{\textbf{Avg.}}} \\
        \cmidrule(lr){2-4}
        & \textbf{w/o Distr.} & \textbf{w/ 2 Distr.} & \textbf{w/ 4 Distr.} & \multicolumn{1}{c}{} \\
        \midrule
        Reason-KE   & \textbf{93.83} & \textbf{89.23} & \textbf{87.53} & \underline{\textbf{90.20}} \\
        \cmidrule{1-5} 
        -w/o acknowledge & 92.77{\scriptsize \textcolor{green}{$\downarrow$}} & 81.00{\scriptsize \textcolor{red}{$\downarrow$}} & 78.67{\scriptsize \textcolor{red}{$\downarrow$}} & \underline{84.15}{\scriptsize \textcolor{red}{$\downarrow$}} \\ 
        -w/o relevance & 88.76{\scriptsize \textcolor{green}{$\downarrow$}} & 86.47{\scriptsize \textcolor{green}{$\downarrow$}} & 80.40{\scriptsize \textcolor{red}{$\downarrow$}} & \underline{87.18}{\scriptsize \textcolor{green}{$\downarrow$}} \\
        -w/o apply & 87.77{\scriptsize \textcolor{red}{$\downarrow$}} & 74.70{\scriptsize \textcolor{red}{$\downarrow\downarrow$}} & 73.67{\scriptsize \textcolor{red}{$\downarrow\downarrow$}} & \underline{78.71}{\scriptsize \textcolor{red}{$\downarrow$}} \\
        -w/o reasoning & 92.17{\scriptsize \textcolor{green}{$\downarrow$}} & 81.77{\scriptsize \textcolor{red}{$\downarrow$}} & 78.43{\scriptsize \textcolor{red}{$\downarrow$}} & \underline{84.12}{\scriptsize \textcolor{red}{$\downarrow$}} \\ 
        -w/o Distr. sample & 93.47{\scriptsize \textcolor{green}{$\downarrow$}} & 88.40{\scriptsize \textcolor{green}{$\downarrow$}} & 86.17{\scriptsize \textcolor{green}{$\downarrow$}} & \underline{89.35}{\scriptsize \textcolor{green}{$\downarrow$}} \\ 
        -only answer & 91.97{\scriptsize \textcolor{green}{$\downarrow$}} & 50.60{\scriptsize \textcolor{red}{$\downarrow\downarrow$}} & 38.37{\scriptsize \textcolor{red}{$\downarrow\downarrow$}} & \underline{60.31}{\scriptsize \textcolor{red}{$\downarrow\downarrow$}} \\
        \midrule
    \end{tabular}}
    \caption{\textbf{Ablation study results under varying distractor settings}, showcasing the impact of different components on the multi-hop accuracy. {\scriptsize\color{red}$\downarrow$} and {\scriptsize\color{green}$\downarrow$} are the same as Table~\ref{table:hops}.}
    \label{table:ablation}
\end{table}

\paragraph{Reason-KE works well for other models and datasets.}
To validate the generalization of our method, we conducted analyses on other models and datasets. 
As shown in table~\ref{table:llama}, we implemented and tested our method on Llama-3-8B-Instruct~\cite{grattafiori2024llama}. The results demonstrate that Reason-KE continues to significantly outperform all baseline methods, confirming its effectiveness is not tied to a specific model architecture.

\begin{table}[t]
    \centering
    \resizebox*{1.0\linewidth}{!}{%
    \begin{tabular}{lcccc}
        \toprule
        \multirow{2}{*}{\bf Method} 
            & \multicolumn{3}{c}{\textbf{Multi-hop acc}} 
            & \multirow{2}{*}{\textit{\textbf{Avg.}}} \\
        \cmidrule(lr){2-4}
        & \textbf{w/o Distr.} & \textbf{w/ 2 Distr.} & \textbf{w/ 4 Distr.} & \multicolumn{1}{c}{} \\
        \midrule
        EditCoT   & 51.26 & 23.13{\scriptsize\color{red}$\downarrow\downarrow$} & 24.20{\scriptsize\color{red}$\downarrow\downarrow$} & \underline{32.87} \\
        RAE   & 85.23 & 80.73{\scriptsize\color{green}$\downarrow$} & 78.96{\scriptsize\color{red}$\downarrow$} & \underline{81.64} \\
        \cmidrule{1-5} 
        Reason-KE   & \textbf{94.37} & \textbf{89.47}{\scriptsize\color{green}$\downarrow$} & \textbf{87.53}{\scriptsize\color{red}$\downarrow$} & \underline{\textbf{90.46}} \\
        \midrule
    \end{tabular}}
    \caption{\textbf{Performance on MQuAKE-CF with Llama-3-8B-Instruct.} All markers follow the same conventions as Table~\ref{table:hops}.}
    \label{table:llama}
\end{table}

Moreover, we evaluated Reason-KE on the DUNE dataset~\cite{dune}, specifically on the Arithmetic, New-Info, and Scientific subsets. These tasks go beyond simple factual triples and involve reasoning with conflicting or novel information. As shown in Table~\ref{tab:dune_performance}, Reason-KE again achieves state-of-the-art performance, proving its applicability to more open-ended and diverse editing scenarios.

\begin{table}[t]
    \centering
    \resizebox*{1.0\linewidth}{!}{%
    \begin{tabular}{llcccc}
        \toprule
        \multirow{2}{*}{\bf Subest} 
        & \multirow{2}{*}{\bf Method} 
            & \multicolumn{3}{c}{\textbf{Acc}} 
            & \multirow{2}{*}{\textit{\textbf{Avg.}}} \\
        \cmidrule(lr){3-5}
        & & \textbf{w/o Distr.} & \textbf{w/ 2 Distr.} & \textbf{w/ 4 Distr.} & \multicolumn{1}{c}{} \\
        \midrule
        \multirow{2}{*}{Arithmetic}         
        & EditCoT & 92.30 & 89.20{\scriptsize\color{green}$\downarrow$} & 90.42{\scriptsize\color{green}$\downarrow$} & \underline{90.64} \\
        & Reason-KE & \textbf{97.46} & \textbf{95.11}{\scriptsize\color{green}$\downarrow$} & \textbf{95.21}{\scriptsize\color{green}$\downarrow$} & \underline{\textbf{95.93}} \\
        \cmidrule(lr){1-6}
        \multirow{2}{*}{New-Info} 
        & EditCoT & 81.20 & 80.10{\scriptsize\color{green}$\downarrow$} & 78.30{\scriptsize\color{green}$\downarrow$} & \underline{79.87} \\
        & Reason-KE & \textbf{84.44} & \textbf{83.35}{\scriptsize\color{green}$\downarrow$} & \textbf{84.06}{\scriptsize\color{green}$\downarrow$} & \underline{\textbf{83.95}} \\
        \cmidrule(lr){1-6}
        \multirow{2}{*}{Scientific}         
        & EditCoT & 81.03 & \textbf{81.23}{\scriptsize\color{green}$\downarrow$} & \textbf{80.70}{\scriptsize\color{green}$\downarrow$} & \underline{80.99} \\
        & Reason-KE & \textbf{82.31} & 80.71{\scriptsize\color{green}$\downarrow$} & 80.59{\scriptsize\color{green}$\downarrow$} & \underline{\textbf{81.20}} \\
        \bottomrule
    \end{tabular}}
    \caption{\textbf{Performance on DUNE Dataset Subsets.} All markers follow the same conventions as Table~\ref{table:hops}.}
    \label{tab:dune_performance}
\end{table}

\paragraph{Does every composition of the training data matter?}
We focus on the importance of the reasoning process in Reason-KE. As shown in Table~\ref{table:ablation}, each component contributes to the reasoning capabilities of the model. Removing any particular segment of the training data disrupts the chain of reasoning and consistently degrades performance.
Notably, completely removing the reasoning process induces over-dependence on the updated facts, leading to a sharp performance decline when facing distractors, with an average performance drop of 29.89\%.
Similarly, excluding distractor-handling samples from the training data also impairs the model's anti-interference ability.

\paragraph{Does Reason-KE require a powerful teacher model?}
A potential concern is the reliance on a powerful teacher model (i.e., DeepSeek-R1) for data generation. To investigate this, we conduct an ablation study by replacing it with a more accessible and less powerful model, Llama-3-8B-Instruct, to generate 1,000 training instances. As shown in Table~\ref{tab:teacher_ablation}, while finetuning on this data leads to a slight performance decrease, the model remains highly competitive and still significantly outperforms the baselines. 
This result confirms that the Reason-KE framework is robust and its success is not contingent on a single proprietary teacher model.

\begin{table}[t]
    \centering
    \resizebox*{1.0\linewidth}{!}{%
    \begin{tabular}{lcccc}
        \toprule
        Method & w/o Distr. & w/ 2 Distr. & w/ 4 Distr. & Avg. \\
        \midrule
        RAE & 78.77 & 71.03{\scriptsize\color{red}$\downarrow$} & 67.87{\scriptsize\color{red}$\downarrow$} & \underline{72.56} \\
        \midrule
        Reason-KE (Llama-3-8B teacher) & 89.50 & 83.26{\scriptsize\color{red}$\downarrow$} & 81.70{\scriptsize\color{red}$\downarrow$} & \underline{84.82} \\
        Reason-KE (DeepSeek-R1 teacher) & \textbf{92.36} & \textbf{87.66}{\scriptsize\color{green}$\downarrow$} & \textbf{86.00}{\scriptsize\color{red}$\downarrow$} & \underline{\textbf{88.67}} \\
        \bottomrule
    \end{tabular}}
    \caption{\textbf{Performance of \texttt{Reason-KE}} trained with data from different teacher models. All markers follow the same conventions as Table~\ref{table:hops}.}
    \label{tab:teacher_ablation}
\end{table}

\begin{figure}[t]
\centering
\resizebox*{1.0\linewidth}{!}{%
  \includegraphics[width=\columnwidth]{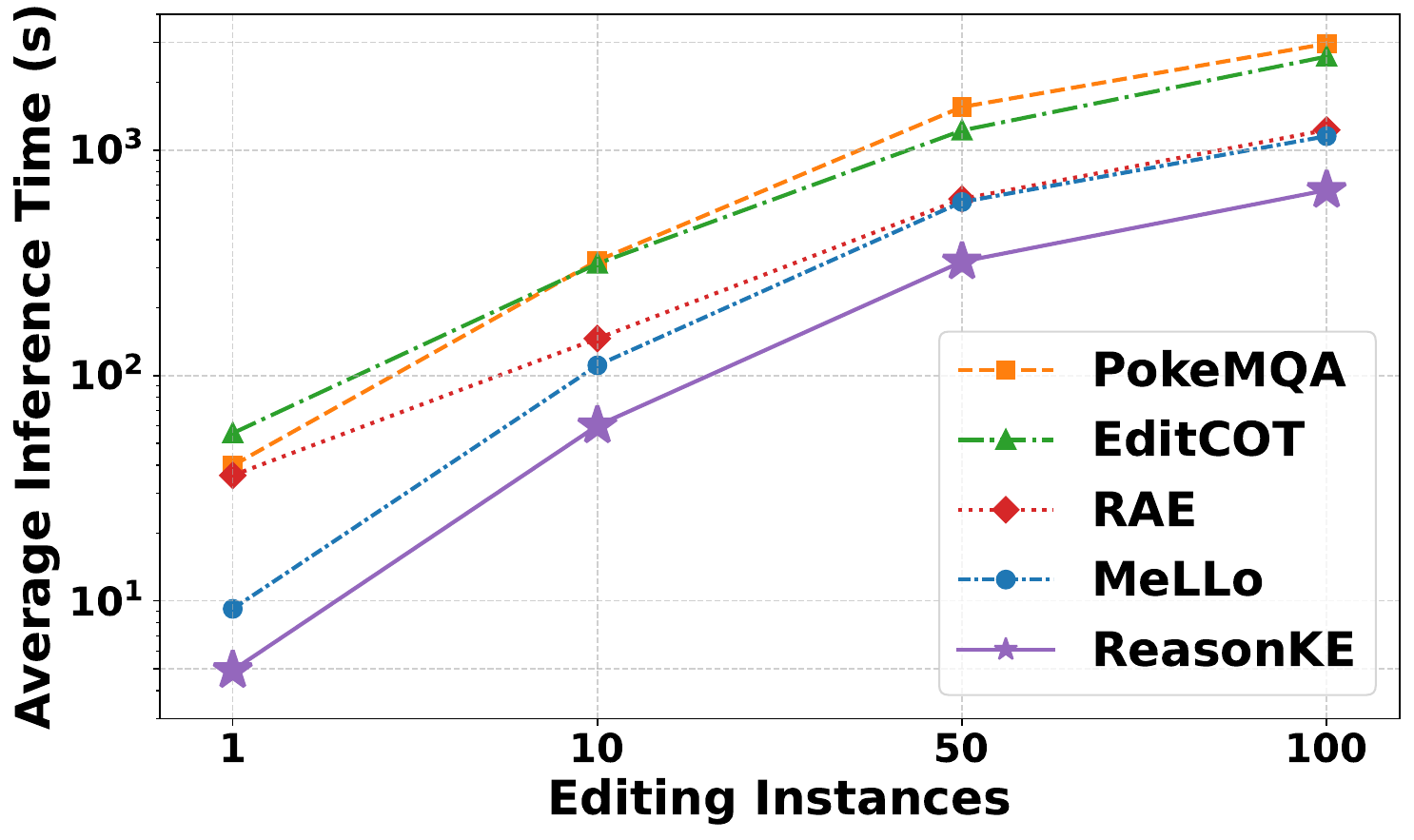}}
  \caption{
  \textbf{Average inference time} for $n$ editing instances, where $n = 1, 10, 50, 100$.}
  \label{fig:infer}
\end{figure}

\paragraph{Reason-KE demonstrates high efficiency in various scenarios.}

We investigate the inference efficiency of Reason-KE and other in-context-learning methods using 100 randomly sampled questions from MQuAKE-CF-3k.
Figure~\ref{fig:infer} illustrates the average time each method needs to edit $n$ cases ($n = 1, 10, 50, 100$), when facing $k$ distractors ($k\in \{0,1,2\}$.).
Due to their high dependence on iterative strategies, most existing methods require more time to reach a final answer, especially in complex scenarios.
In contrast, Reason-KE only requires an editing prompt to perform reasoning, achieving a significant efficiency improvement.

\section{Conclusion}
\label{sec:conclusion}
We present Reason-KE, a streamlined end-to-end framework that guides an LLM through four explicit reasoning stages—acknowledgment, relevance filtering, selective application, and final inference—to perform multi-hop knowledge edits in one pass. On MQuAKE-CF-3K, Reason-KE boosts Qwen2.5-7B’s accuracy to 90.2\% (↑17.6 pp) while limiting drops to 6.3\% with four distractors and under 1\% when answers leak. Ablations confirm that each reasoning stage is critical to robustness, and inference is markedly faster than iterative baselines. Reason-KE offers a simple, reliable, and efficient solution for updating LLM world knowledge.

\section*{Limitations}
While our work presents promising results, several limitations should be noted. First, due to computational constraints, we validate Reason-KE on models with up to 7B parameters. Evaluating larger models, especially those exceeding 70B parameters, could provide more comprehensive insights. 
Second, although our method performs well in multi-hop settings, its potential in other domains like finance or law remains unexplored. Future research will focus on scaling Reason-KE and extending its application to additional fields beyond the knowledge triple setup.

\section*{Ethics and Reproducibility Statements}
\paragraph{Ethics}  
We take ethical considerations seriously and strictly adhere to the ACL Ethics Policy. All datasets used in this work are publicly available and widely adopted by the research community. Our methods focus on enhancing the multi-hop QA knowledge editing capabilities of large language models without introducing harmful biases or unethical content. We ensure that all experiments are conducted in compliance with ethical guidelines, emphasizing fairness and transparency in model deployment.

\paragraph{Reproducibility}  
In this paper, we discuss the detailed experimental setup, including training hyperparameters, baseline implementations, and statistical descriptions. More importantly, in addition to the code repository provided in the abstract, we have uploaded our data (\url{https://huggingface.co/datasets/YukinoKaren/Reason-KE-train-data}) and model (\url{https://huggingface.co/YukinoKaren/Reason-KE}) to Hugging Face to facilitate reproduction of the experimental results in this work. 

\normalem
\bibliography{custom}

\appendix
\onecolumn

\section{Details of Dataset Construction}
\label{sec:Dataset_sample}
For each item in MQuake-CF, we generate a corresponding reasoning process. In total, we construct 9,218 data entries for the training set.
Here, we present the detailed prompts used for sample generation. Specifically, we employ the following prompts to guide LLMs to produce the reasoning process.

\begin{tcolorbox}
[colback=lightgray!20,colframe=darkgray!80,title= Reasoning Process Generation Prompt]
\label{tab:quality_prompt}
Please provide a reasoning process based on my following tasks and corresponding answers. Your answer must strictly follow the steps of my example.
\newline
\newline
[Task]:Please acknowledge the updated information provided below and respond to the subsequent query.

[Updated Information]:Roblin Park is located in New South Wales.

[Query]:What is the capital city of the state where Roblin Park is located?

[Answer]:Sydney

[Reasoning Process]

1.Acknowledge Updated Information: The updated information states that Roblin Park is located in New South Wales.

2.Determine Relevance: The query asks for the capital of the state where Roblin Park is located. Since the updated information explicitly provides the state (New South Wales), it is directly relevant to answering the question.

3.Apply Updated Information or Ignore: Apply Roblin park's new location.

4.Reasoning: Roblin Park lies within the state of New South Wales. The capital of New South Wales is Sydney. Therefore, the capital of the state where Roblin Park is located is Sydney

[Answer]: Sydney
\newline
\newline
[Task]:Please acknowledge the updated information provided below and respond to the subsequent query.
\newline
[Updated Information]: {\color{red}\texttt{<updated\_information>}}
\newline
[Query]: {\color{red}\texttt{<query>}}
\newline
[Answer]: {\color{red}\texttt{<answer>}}

\end{tcolorbox}

\section{Details of Experimental Setup}


\subsection{Details of Baselines}
\label{sec:details_baselines}
We compare Reason-KE with parameter modification method and current In-Context Editing methods:
\paragraph{ROME~\cite{rome}}employs causal mediation analysis to pinpoint the target area for editing and subsequently updates the parameters of the feed-forward network. In our implementation, we utilize EasyEdit~\cite{easyedit} with its default settings.
\paragraph{MeLLo~\cite{mquake}}employs the plan-and-solve strategy to perform in-context editing. It first decomposes the problem into sub-questions and uses retrieval. Following the official setting, the prompts were adapted to Instruct Models, and the maximum number of retrieval rounds was fixed at four.
\paragraph{PokeMQA\cite{pokemqa}}building upon Mello, enhances question understanding by prompting large language models (LLMs) to decompose knowledge-augmented multi-hop questions. We adhere to the official settings, allowing up to five rounds of interaction and utilizing their pre-trained Scope-Detector. 
\paragraph{EditCoT~\cite{EDITCOT}}employs an iterative CoT approach. It first generates an initial CoT based on user input and the direct answer. A dedicated CoT editor then revises this reasoning trace, injecting newly retrieved knowledge to resolve any conflicts or gaps. Once the CoT has been updated, the language model is prompted to reason along the refined path and generate the final answer. Following the official setting, the maximum number of retrieval rounds was fixed at four.
\paragraph{RAE~\cite{rae}}constructs retrieval-oriented knowledge graphs and uses the model to optimize graph retrieval and pruning.

\subsection{Details of Datasets}
\label{sec:appendix_Datasets}
Table~\ref{datasetstatistic} shows the statistics of the MQUAKE-CF-3k datasets, which contain 3000 data points.

\begin{table}[h]
\centering
\begin{tabular}{llcccc}
\toprule
 \textbf{Datasets} & \textbf{\#Edits} & \multicolumn{1}{l}{\textbf{2-hop}} & \multicolumn{1}{l}{\textbf{3-hop}} & \multicolumn{1}{l}{\textbf{4-hop}} & \multicolumn{1}{l}{\textbf{Total}} \\ \hline
                                 & 1       & 513  & 356  & 224  & 1093 \\
                                 & 2       & 487  & 334  & 246  & 1067 \\
\multicolumn{1}{c}{MQUAKE-CF-3K} & 3       & -    & 310  & 262  & 572  \\
                                 & 4       & -    & -    & 268  & 268  \\
                                 & All     & 1000 & 1000 & 1000 & 3000 \\ 
\bottomrule
\end{tabular}
\caption{Statistics of MQuAKE-CF-3K datasets.}
\label{datasetstatistic}
\end{table}

\subsection{Implementation Details}
\label{sec:appendix_implementation}
All experiments were executed on 4 NVIDIA A100 GPUs (80G). We implement our Reason-KE method by supervised fine-tuning on Qwen2.5-7B-Instruct~\cite{qwen25}, with the training duration approximately lasting 100 to 120 minutes. Hyperparameters of our Reason-KE method are in Table \ref{tab:hyperparameters}.

\begin{table}[!h]
\centering
{\begin{tabular}{lc}
\toprule
\textbf{Hyperparameter} & \textbf{Standard-FT} \\ 
\midrule
Epoches                 & 5  \\
Warmup ratio            & 0.05 \\
Optimizer               & AdamW \\
Learning rate           & 1e-5                     \\
Scheduler               & cosine                 \\  
Weight decay            & 1e-4 \\
Block size              & 32768 
\\ \bottomrule
\end{tabular}}
\caption{\label{tab:hyperparameters}
\textbf{Hyper-parameters} for training our Reason-KE.}
\end{table}

\subsection{Details of Distractors Selection.}
\label{sec:selection}
We employ Contriever~\cite{Contriever} for retrieval and TopK~\cite{toks} as the baseline selection method.
Specifically, for each edit fact, we retrieve from the pre-edit corpus and match with the Top-k corresponding post-edit facts, where $ k \in{0,1,2}$.

\section{Used Scientific Artifacts}
\label{sec:Scientific}
Below is a list of the scientific artifacts used in our work. Our use of these existing artifacts is consistent with their intended use.

\begin{itemize} 
    
    \item \textit{DeepSpeed (Apache-2.0 license)}~\footnote{ \url{https://github.com/deepspeedai/DeepSpeed}}, a deep learning optimization library that enhances the efficiency of large language model training.
    
    \item \textit{Transformers (Apache-2.0 license)}~\footnote{ \url{https://github.com/huggingface/transformers}}, a framework offering state-of-the-art pretrained models for NLP tasks.

    \item \textit{trl (Apache-2.0 license)}~\footnote{ \url{https://github.com/huggingface/trl}}, a library that combines reinforcement learning (RL) with transformer models.

\end{itemize}

\end{document}